\documentclass[sn-mathphys]{sn-jnl}

\usepackage{amsmath,amssymb,amsfonts}
\usepackage{graphicx}
\usepackage{textcomp}
\usepackage{xcolor}
\usepackage{array}
\usepackage{textcomp}
\usepackage{stfloats}
\usepackage{url}
\usepackage{verbatim}
\usepackage{bbding}
\usepackage{booktabs}
\usepackage{float}
\usepackage{subfigure}
\usepackage{setspace}
\usepackage{soul}
\usepackage{color}
\usepackage{amssymb}
\usepackage{balance}

\definecolor{c1}{HTML}{006E5F}
\definecolor{c5}{HTML}{ED9F54}

\jyear{2022}%

\theoremstyle{thmstyleone}%
\theoremstyle{thmstyletwo}%

\theoremstyle{thmstylethree}%

\begin{document}

\title[Heterogeneous Tri-stream Clustering Network]{Heterogeneous Tri-stream Clustering Network}

\author[1]{\mbox{Xiaozhi Deng}}\email{dengxiaozhi45@gmail.com}

\author*[1,2]{\mbox{Dong Huang}}\email{huangdonghere@gmail.com}

\author[3]{\mbox{Chang-Dong Wang}}\email{changdongwang@hotmail.com}

\affil*[1]{\orgdiv{College of Mathematics and Informatics}, \orgname{South China Agricultural University}, \orgaddress{\city{Guangzhou}, \country{China}}}

\affil[2]{\orgname{Key Laboratory of Smart Agricultural Technology in Tropical South China, Ministry of Agriculture and Rural Affairs}, \orgaddress{\country{China}}}

\affil[3]{\orgdiv{School of Computer Science and Engineering}, \orgname{Sun Yat-sen University}, \orgaddress{\city{Guangzhou}, \country{China}}}


\abstract{Contrastive deep clustering has recently gained significant attention with its ability of joint contrastive learning and clustering via deep neural networks. Despite the rapid progress, previous works mostly require both positive and negative sample pairs for contrastive clustering, which rely on a relative large batch-size. Moreover, they typically adopt a two-stream architecture with two augmented views, which overlook the possibility and potential benefits of multi-stream architectures (especially with heterogeneous or hybrid networks). In light of this, this paper presents a new end-to-end deep clustering approach termed Heterogeneous Tri-stream Clustering Network (HTCN). The tri-stream architecture in HTCN consists of three main components, including two weight-sharing online networks and a target network, where the parameters of the target network are the exponential moving average of that of the online networks. Notably, the two online networks are trained by simultaneously (i) predicting the instance representations of the target network and (ii) enforcing the consistency between the cluster representations of the target network and that of the two online networks. Experimental results  on four challenging image datasets demonstrate the superiority of HTCN over the state-of-the-art deep clustering approaches. The code is available at \url{https://github.com/dengxiaozhi/HTCN}.~~~~~~~~~}




\maketitle

\section{Introduction}\label{sec1}

Data clustering is the process of grouping data samples into multiple clusters in an unsupervised manner, which is a fundamental task in a variety of applications \cite{Frey2007,Huang2018,huang20_tkde}. The traditional clustering algorithms typically focus on some low-level information and lack the representation learning ability, which may lead to sub-optimal performance when dealing with some complex high-dimensional data like images.

In recent years, the deep learning has gained tremendous progress \cite{yu2019hierarchical,zhang2018local,zhang2022semisupervised}, which has also been exploited for tackling the clustering task,  giving rise to the rapid development of the deep clustering algorithms \cite{xie2016unsupervised,guo2017improved,wang18-npl,ji2019invariant,Chen2022}. For example, Xie et al. \cite{xie2016unsupervised} presented a deep clustering method called Deep Embedded Clustering (DEC), which simultaneously learns representations and cluster assignments with an objective loss based on Kullback-Leibler (KL) divergence. Guo et al. \cite{guo2017improved} extended DEC by incorporating the reconstruction loss (via autoencoder) to preserve local structures. Ji et al. \cite{ji2019invariant} sought to learn invariant information of data by maximizing the mutual information between paired samples.
More recently, the contrastive learning has emerged as a promising technique for exploiting sample-wise (or augmentation-wise) contrastiveness to improve the deep clustering performance. Van Gansbeke et al. \cite{van2020scan} presented the Semantic Clustering by Adopting Nearest neighbors (SCAN) method, which first adopts contrastive learning to learn discriminant features and then performs semantic clustering with the $K$-nearest neighbors exploited. Dang et al. \cite{dang2021nearest} matched local-level and global-level nearest neighbors to further improve clustering performance.
Li et al. \cite{li2021contrastive} presented the Contrastive Clustering (CC) method to perform feature learning and clustering with simultaneous instance-level and cluster-level contrastive learning.

Despite significant success, these contrastive deep clustering methods \cite{van2020scan,dang2021nearest,li2021contrastive} are mostly faced with two limitations. On the one hand, they typically requires both positive sample pairs and negative sample pairs during their contrastive learning process, which rely on a relatively large batch-size (for sufficient negative pairs) and may bring in a heavier computational burden. On the other hand, these prior works generally adopt a two-stream architecture (with two weight-sharing augmented views), which neglect the possibility of going beyond the two-stream architecture to utilize three or even more streams of networks (with heterogeneous or hybrid structures). {Recently Grill et al. \cite{grill2020bootstrap} presented the Bootstrap Your Own Latent (BYOL) method, which adopts an asymmetric two-stream architecture (with an online network and a target network) and conducts the contrastive learning without negative pairs, where the online network is trained by predicting the feature representations of the target network.}  Though the requirement for negative sample pairs is remedied, BYOL still complies with the two-stream architecture and also lacks the ability of directly learning the clustering structure. It remains a challenging problem how to incorporate contrastive learning into multiple streams of heterogeneous networks while alleviating the dependence on negative sample pairs for strengthened deep clustering performance.

{In light of this, this paper presents a novel deep clustering approach termed Heterogeneous Tri-stream Clustering Network (HTCN), which leverages three streams of heterogeneous networks for simultaneous cluster-level and instance-level contrastive learning without requiring negative sample pairs (as illustrated in Fig.~\ref{fig1}).} Inspired by BYOL \cite{grill2020bootstrap}, we design a novel tri-stream architecture with three augmented views, corresponding to two online networks and a target network, respectively. {Note that the online network and the target network are heterogeneous, which differ from each other in the network structure and the updating mechanism.
The two online networks share the same parameters, while the parameters of the target network are the exponential moving average of that of the online networks. Here, the exponential moving average is a type of moving average that places a greater weight and significance on the most recent data samples \cite{grill2020bootstrap}.} Each online network is associated with an instance predictor and a cluster predictor, which produce the instance-level representations and the cluster-level representations, respectively. Different from the online networks, the target network utilizes a cluster predictor to generate the cluster-level representations while producing the instance-level representations by the projector directly.
{The incorporation of an instance predictor in the online networks is meant to prevent the potential collapse where the networks produce the same feature representations for most samples.} Then we train the two online networks by (i) predicting the target network's representation of the same image via the mean squared error (MSE) loss (for the instance-level contrastive learning) and (ii) enforcing the consistency between the predicted cluster distributions of the two online networks and that of the target network via the information noise contrastive estimation (InfoNCE) \cite{oord2018representation} loss (for the cluster-level contrastive learning).
Experiments conducted on four image datasets demonstrate the superiority of our approach over the state-of-the-art deep clustering approaches.

For clarity, the contributions of this work are summarized below.
\begin{itemize}
  \item A heterogeneous tri-stream architecture is designed, where two online networks and a target network are jointly leveraged for instance-level and cluster-level contrastive learning.
  \item A novel deep clustering approach termed HTCN is proposed, which utilizes three augmented views for contrastive learning without requiring negative sample pairs.
  \item Experimental results on four image datasets confirm the advantegeous clustering performance of our HTCN approach over the state-of-the-art deep clustering approaches.
\end{itemize}

The rest of the paper is organized as follows. The related works on deep clustering and self-supervised learning are reviewed in Section~\ref{sec2}. The proposed HTCN framework is described in Section~\ref{sec3}. The experiments are reported in Section~\ref{sec4}. Finally, Section~\ref{sec5} concludes the paper.

\section{Related Work}\label{sec2}

In this section, we will introduce the related works on deep clustering and self-supervised learning.

\subsection{Deep Clustering}

Traditional clustering methods such as $K$-means \cite{macqueen1967some} and spectral clustering (SC) \cite{zelnik2004self} have achieved promising results in handling low-dimensional data, but they may result in sub-optimal performance when faced with high-dimensional data (e.g., images and videos) due to the lack of the representation learning ability. To address this, the deep learning based clustering methods, referred to as the deep clustering methods, have recently achieved significant success, which leverage the power of feature learning of deep neural networks for the clustering task 
\cite{li2021contrastive,huang2020deep,wang18-npl,ji2019invariant,guo2019adaptive,caron2018deep,guo2017improved,xie2016unsupervised,yang2016joint,zhu20_npl,dang2021nearest,park2021improving,van2020scan,han2020mitigating,zhong2020deep,zhu21_npl}.

Previous deep clustering methods can be divided into two main categories, namely, the one-stage methods 
and the two-stage methods. The goal of the one-stage approach is to perform feature representation learning and clustering assignment simultaneously. Xie et al. \cite{xie2016unsupervised} proposed a Deep Embedding Clustering (DEC) method, which jointly optimizes feature learning and clustering with a KL-divergence loss. Caron et al. \cite{caron2018deep} iteratively clustered the learned features with $K$-means and regarded the cluster assignments as supervisory signals to optimize the network. Li et al.\cite{li2021contrastive} presented a Contrastive Clustering (CC) method that performs contrastive learning at instance-level and cluster-level for deep clustering. Besides the one-stage methods, some researchers have also made considerable efforts to the two-stage clustering methods. Van Gansbeke et al.\cite{van2020scan} proposed a two-stage clustering method called Semantic Clustering by Adopting Nearest neighbors (SCAN), which first learns the semantic features via contrastive learning and then utilizes the features for clustering in the next stage. To extend SCAN, Dang et al. \cite{dang2021nearest} designed a Nearest Neighbor Matching (NNM) method, which selects both local and global nearest neighbors to optimize the network, where the neighbors are forced to be close to each other.

\subsection{Self-supervised Learning}

Self-supervised learning has recently emerged as a powerful technique with the ability to learn representation from raw data without human supervision, in which the contrastive learning methods \cite{wu2018unsupervised,misra2020self,he2020momentum,chen2020simple} have been a representative and promising category.

The goal of contrastive learning is to minimize the distance between positive sample pairs while maximizing the distance between negative sample pairs in a self-supervised manner, where positive pairs and negative pairs are defined through data augmentations. In particular, some researchers maintained a memory bank \cite{wu2018unsupervised,misra2020self} that contains large amounts of representations of negative samples to achieve high performance. However, these methods that utilize memory banks to store and update representations may be computationally expensive. To address the problems with memory banks, He et al. \cite{he2020momentum} proposed a Momentum Contrast (MoCo) method that trains an encoder by the momentum update mechanism maintaining a long queue of negative examples. Following the MoCo method, Chen et al. \cite{chen2020simple} proposed a Simple framework for Contrastive LeaRning (SimCLR) method which carefully designs the strategy of data augmentation and a non-linear transformation head.
In addition, the clustering based methods \cite{caron2020unsupervised,li2020prototypical} adopt a clustering approach to group similar features together, which address the issue that every sample is considered as a discrete class in previous works. More recently, some self-supervised learning methods that only rely on positive pairs and directly predict the output of one augmented view from another augmented view \cite{grill2020bootstrap,chen2021exploring,li2022tribyol} have been developed, among which a representative method is the BYOL method \cite{grill2020bootstrap}. The BYOL method \cite{grill2020bootstrap} adopts an asymmetric two-stream architecture, which, however, lacks the ability to learn the clustering structure directly and also overlooks the opportunities and potential benefits of going beyond the two-stream architecture to three or more streams of networks (even with heterogeneous or hybrid structures) to further enhance the contrastive learning and clustering performance.

\section{Proposed Framework}\label{sec3}


\subsection{Framework Overview}
\label{sec:framework_overview}

\begin{figure*}[!t]
\centering
\includegraphics[width=1\textwidth]{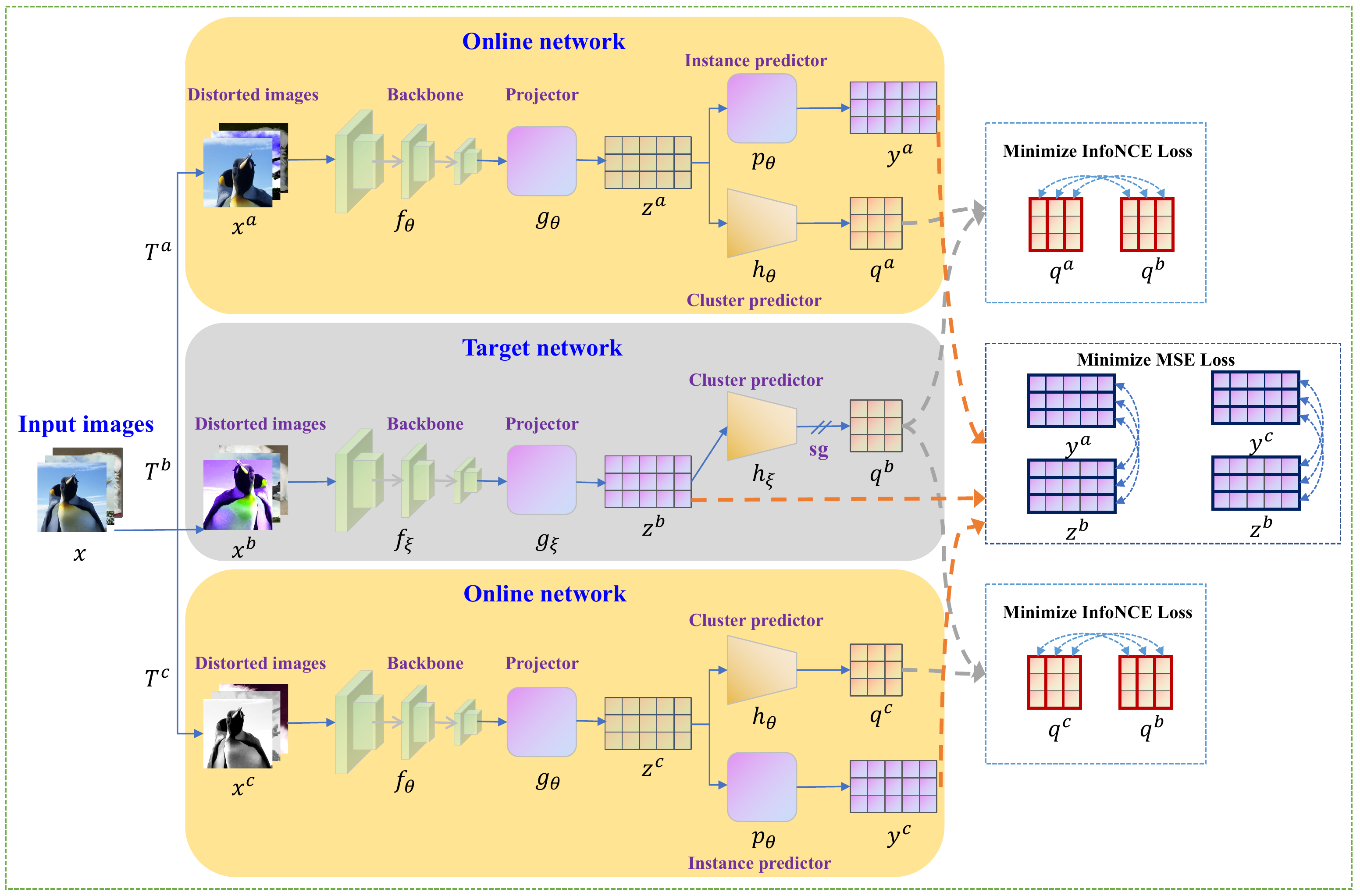}
\caption{Illustration of the proposed HTCN framework. The tri-stream network consists of two weight-sharing online networks and a target network, where the parameters of the target network is an exponential moving average of that of the online networks. Instance predictors and cluster predictors are incorporated in the three networks, after which the MSE loss and the InfoNCE loss are ultilized for instance-level contrastive learning and cluster-level contrastive learning, respectively. The network architecture can be trained in an end-to-end manner, where the final clustering is obtained via the cluster predictor of the target network.}
\label{fig1}
\end{figure*}

This paper presents a heterogeneous  tri-stream network architecture termed HTCN for contrastive deep clustering (as illustrated in Fig.~\ref{fig1}), which goes beyond the traditional two-stream architecture to explore the constrastive network in a multi-stream manner. Also, HTCN doesn't require negative sample pairs, which makes it more resilient to different batch-size. Specifically, HTCN consists of three main components, including two online networks and a target network. The online networks and the target network are respectively parameterized by different sets of weights, where the parameters of the target network are an exponential moving average of that of the online networks.


Given a batch of $N$ images, we perform three types of augmentations on each image, denoted as $x_{i}$ with $i \in[1, N]$, to generate $3\cdot N$ augmented (or distorted) images, denoted as $\{x_{1}^{a}, \ldots, x_{N}^{a}, x_{1}^{b}, \ldots, x_{N}^{b}, x_{1}^{c}, \ldots, x_{N}^{c}\}$. The backbones (i.e., $f_{\theta}$ and $f_{\xi}$) and projectors (i.e., $g_{\theta}$ and $g_{\xi}$) are adopted to extract features from the distorted images via $z_{i}^{a} = g_{\theta}(f_{\theta}(x_{i}^{a}))$, $z_{i}^{b} = g_{\xi}(f_{\xi}(x_{i}^{b}))$ and $z_{i}^{c} = g_{\theta}(f_{\theta}(x_{i}^{c}))$. Then the instance predictors transform $z^{a}_{i}$ and $z^{c}_{i}$ to $y^{a}_{i}$ and $y^{c}_{i}$, respectively, while the cluster predictors  transform $z^{a}_{i}$, $z^{b}_{i}$ and $z^{c}_{i}$ to $\tilde{q}^{a}_{i}$, $\tilde{q}^{b}_{i}$ and $\tilde{q}^{c}_{i}$, respectively. Note that, similar to the asymmetric architecture of BYOL, the target network is not associated with an instance predictor, and the representations generated by its projector are used to guide the instance-level learning of the two online networks. {The row space of the feature matrix learned by the projector or the instance predictor is expressed as the instance-level representations, while the column space of the feature matrix learned by the cluster predictor is expressed as the cluster-level representations. The instance-level representations are utilized to enforce the instance-level contrastive learning with an MSE loss optimized, while the cluster-level representations are utilized to enforce the cluster-level contrastive learning with an InfoNCE loss optimized.} Finally, the instance-level and cluster-level contrastive losses are simultaneously utilized to optimize the tri-stream network.

\subsection{Instance-level Contrastiveness}
\label{sec:framework consistency}

Our HTCN approach simultaneously performs feature learning and clustering without requiring negative sample pairs. In instance-level contrastive learning, we aim to train the two online networks by predicting the instance representations of target network. Specifically, let $y^{a}_{i}$ and $z^{b}_{i}$ be the instance representations of $x_i$ in the first online network and the target network, respectively. The instance-level contrastive loss between them is defined as
\begin{equation}
	\mathcal{L}_{a, b, i} =\|\overline{y^{a}_{i}}-\overline{z^{b}_{i}}\|_{2}^{2}=2-2 \cdot \frac{\langle y^{a}_{i}, z^{b}_{i}\rangle}{\|y^{a}_{i}\|_{2} \cdot\|z^{b}_{i}\|_{2}} ,
\end{equation}
where $\overline{y^{a}_{i}}$ and $\overline{z^{b}_{i}}$ are the normalized representations. Thus the loss between the first and second views can be expressed as
\begin{equation}\label{}
	\mathcal{L}_{a, b}=\frac{1}{N} \sum_{i=1}^{N}\mathcal{L}_{a, b, i}
\end{equation}
Similar to BYOL \cite{grill2020bootstrap}, the exchange of the online and target views is performed during each training step. Also, we utilize another online network to predict the representations produced by the target network, whose loss is defined as
\begin{align}
	\mathcal{L}_{b, c, i} =\|\overline{y^{c}_{i}}-\overline{z^{b}_{i}}\|_{2}^{2}=&2-2 \cdot \frac{\langle y^{c}_{i}, z^{b}_{i}\rangle}{\|y^{c}_{i}\|_{2} \cdot\|z^{b}_{i}\|_{2}} , i \in[1, N],\\
	\mathcal{L}_{b, c}=&\frac{1}{N} \sum_{i=1}^{N}\mathcal{L}_{b, c, i},
\end{align}
Therefore, the instance-level contrastive loss among the three streams of networks is defined as
\begin{equation}
	\mathcal{L}_{instance} = \mathcal{L}_{a, b} + \mathcal{L}_{b, c} .
\end{equation}
\subsection{Cluster-level Contrastiveness}
\label{sec:framework contrastiveness}

The cluster predictor maps the representations produced by the projector to $M$-dimensional probability vectors, where $M$ is the number of clusters. These probability vectors, whose $i$-th element denotes how likely the image belongs to the $i$-th cluster, can be interpreted as the soft label.
Let $q^{a},q^{b},q^{c}\in\mathbb{R}^{N\times M}$ be the feature matrices produced by the cluster predictors of the three networks, respectively. Each column of the feature matrix denotes an $N$-dimensional cluster representation, denoted as $q^{k}_{i}$, while the each row denotes a $M$-dimensional probability vector, denoted as $\tilde{q}^{k}_{i}$ (for $k\in\{a,b,c\}$).


For a cluster representation $q^{a}_{i}$, we regard $q^{a}_{i}$ and $q^{b}_{i}$ as a positive cluster pair, and the other $2\cdot M - 2$ pairs (in the first and second views) as the negative cluster pairs. The pair-wise similarity is defined as
\begin{align}\label{}
	s(q^{a}_{i}, q^{b}_{j})=\frac{\langle q^{a}_{i}, q^{b}_{j}\rangle}{\|q^{a}_{i}\|\|q^{b}_{j}\|},~~i,j \in[1, M]
\end{align}
Then the InfoNCE loss for $q^{a}_{i}$ is computed by
\begin{equation}\label{}
	\ell_{i}^{a}=-\log \frac{\exp (s(q^{a}_{i}, q^{b}_{j}) / \tau)}{\sum_{j=1}^{M}[\exp (s(q^{a}_{i}, q^{a}_{j}) / \tau)+\exp (s(q^{a}_{i}, q^{b}_{j}) / \tau)]} ,
\end{equation}
where $\tau$ is the temperature parameter. After traversing all cluster representations, the cluster-level contrastive loss between the first and second augmented views can be obtained as
\begin{align}\label{}
	\hat{\mathcal{L}}_{a, b}=\frac{1}{2 M} \sum_{i=1}^{M}(\ell_{i}^{a}+\ell_{i}^{b})-H(Q),
\end{align}
where $H(Q)$ is the entropy of the cluster-assignment probability, which helps to avoid a degenerate solution that most images fall into the same cluster and is computed as
\begin{align}\label{}
	H(Q)=&-\sum_{i=1}^{M}[P(q^{a}_{i}) \log P(q^{a}_{i})+P(q^{b}_{i}) \log P(q^{b}_{i})],\\
	P(q_{i}^{k})=&\sum_{j=1}^{N} \frac{q_{j i}^{k}} {\|q\|_{1}},~~k \in\{a, b\}
\end{align}
For each batch of images, a view pair is formed between each online network and the target network, leading to a total of two view pairs for the cluster-level contrastive learning. Therefore, the cluster-level contrastive loss can be defined as

\begin{equation}\label{}
	\mathcal{L}_{cluster}=\hat{\mathcal{L}}_{a, b}+\hat{\mathcal{L}}_{b, c}.
\end{equation}

\subsection{Overall Loss Function}
\label{sec:framework loss}

The tri-stream network of HTCN is trained by simultaneously considering the instance-level contrastiveness and the cluster-level contrastiveness. The overall loss function is defined as
\begin{equation}
	\mathcal{L} = \mathcal{L}_{instance} + \mathcal{L}_{cluster} .
\end{equation}


At each training step, we optimize the overall loss function w.r.t. the online networks' parameters $\theta$ only, but not the target network's parameters $\xi$. The parameters of the target is updated as an exponential moving average of that of the online networks. That is
\begin{equation}
	\theta \leftarrow \operatorname{optimizer}(\theta, \nabla_{\theta} \mathcal{L}, \eta),
\end{equation}
\begin{equation}
	\xi \leftarrow \alpha\xi + (1-\alpha)\theta.
\end{equation}
where $\eta$ is the learning rate and $\alpha$ is the momentum coefficient. After the training, we only keep the target network to perform clustering, which can be obtained in the cluster predictor.

\subsection{Implementation Details}
\label{sec:implement_details}

In HTCN, we use the ResNet34 \cite{he2016deep} as the backbone. The projectors and the instance predictors have the same network structure, each of which is a multi-layer perceptron (MLP) with 256-dimensional output units. Each of the cluster predictors is a two-layer MLP, whose output dimension is equal to the desired number of clusters.

Three augmented (or distorted) views are generated by applying a family of transformations to each input image. Five types of augmentations are utilized, including ResizedCrop, HorizontalFlip, ColorJitter, Grayscale and GaussianBlur \cite{li2021contrastive}. {As each transformation has a probability of being adopted, the distortions of the three streams can thus be randomly decided.}
During optimization, we use the Adam optimizer and train the model for 1000 epochs. The learning rate is set to 0.0003.The batch size is set to 128.

\begin{table}[!t]\vskip 0.12 in
	\renewcommand\arraystretch{1}
	\centering
	\caption{Description of the benchmark image datasets.}
	\label{tab1}
	\begin{tabular}{p{3.3cm}<{\centering}p{2cm}<{\centering}p{2cm}<{\centering}}
		\toprule
		Dataset       & \#Images  & \#Classes \\ \midrule
		CIFAR-100     & 60,000  & 20       \\
		ImageNet-10   & 13,000  & 10       \\
		ImageNet-Dogs & 19,500  & 15       \\
		Tiny-ImageNet & 100,000 & 200      \\ \bottomrule
	\end{tabular}
\end{table}

\begin{figure}[!t]
	\centering
	\subfigure[CIFAR-100]{
		\includegraphics[width=0.4843\textwidth]{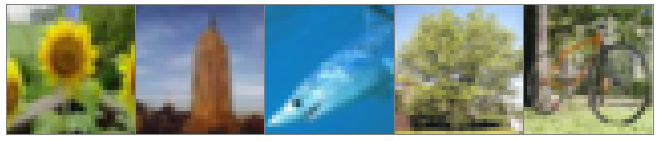}}
	\subfigure[ImageNet-10]{
		\includegraphics[width=0.4843\textwidth]{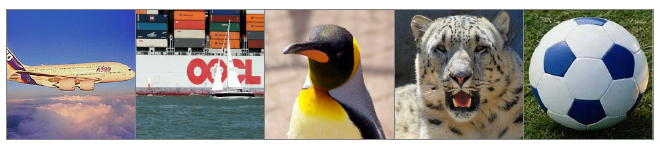}}
	\subfigure[ImageNet-Dogs]{
		\includegraphics[width=0.4843\textwidth]{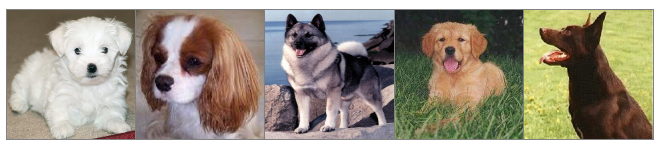}}
	\subfigure[Tiny-ImageNet]{
		\includegraphics[width=0.4843\textwidth]{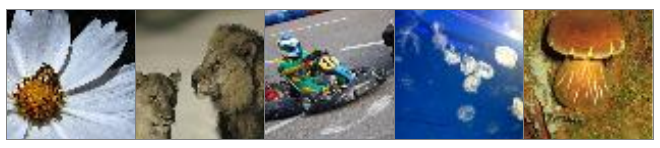}}
	\caption{Some examples of the four image datasets.}
	\label{fig3}\vskip 0.1 in
\end{figure}

\section{Experiments}\label{sec4}


\subsection{Datasets and Evaluation Metrics}

The experiments are conducted on four widely-used image datasets, namely, CIFAR-100 \cite{krizhevsky2009learning}, ImageNet-10 \cite{chang2017deep}, ImageNet-Dogs \cite{chang2017deep}, and Tiny-ImageNet \cite{le2015tiny}. The statistics of these benchmark datasets are given in Table~\ref{tab1}, and some sample images of these datasets are illustrated in Fig.~\ref{fig3}.

To compare the clustering results of different clustering methods, three evaluation metrics are adopted, including  normalized mutual information (NMI) \cite{Huang2021_TSMC}, clustering accuracy (ACC) \cite{Huang2021}, and adjusted rand index (ARI) \cite{liang22_tnnls}. 

\subsection{Comparison with State-of-the-Art}

\begin{table*}[]
	\renewcommand\arraystretch{1}
	\centering
	\caption{The NMI(\%) scores by different clustering methods (The best score in each column is in \textbf{bold}).}
	\label{tab2.1}
	\begin{tabular}{|p{2.04cm}<{\centering}|p{1.8cm}<{\centering}|p{2cm}<{\centering}|p{2.36cm}<{\centering}|p{2.33cm}<{\centering}|}
		\hline
		Dataset      & CIFAR-100 & ImageNet-10 & ImageNet-Dogs & Tiny-ImageNet \\ \hline
		$K$-means \cite{macqueen1967some}      & 8.4             & 11.9            & 5.5             & 6.5            \\ \hline
		SC \cite{zelnik2004self}               & 9.0             & 15.1            & 3.8             & 6.3            \\ \hline
		AC \cite{gowda1978agglomerative}       & 9.8             & 13.8            & 3.7             & 6.9            \\ \hline
		NMF \cite{cai2009locality}             & 7.9             & 13.2            & 4.4             & 7.2            \\ \hline
		AE \cite{bengio2006greedy}             & 10.0            & 21.0            & 10.4            & 13.1           \\ \hline
		DAE \cite{vincent2010stacked}          & 11.1            & 20.6            & 10.4            & 12.7           \\ \hline
		DCGAN \cite{radford2015unsupervised}   & 12.0            & 22.5            & 12.1            & 13.5           \\ \hline
		DeCNN \cite{zeiler2010deconvolutional} & 9.2             & 18.6            & 9.8             & 11.1           \\ \hline
		VAE \cite{kingma2013auto}              & 10.8            & 19.3            & 10.7            & 11.3           \\ \hline
		JULE \cite{yang2016joint}              & 10.3            & 17.5            & 5.4             & 10.2           \\ \hline
		DEC \cite{xie2016unsupervised}         & 13.6            & 28.2            & 12.2            & 11.5           \\ \hline
		DAC \cite{chang2017deep}               & 18.5            & 39.4            & 21.9            & 19.0           \\ \hline
		DCCM \cite{wu2019deep}                 & 28.5            & 60.8            & 32.1            & 22.4           \\ \hline
		GATC \cite{niu2020gatcluster}    & 28.5            & 59.4            & 28.1            & -              \\ \hline
		PICA \cite{huang2020deep}              & 31.0            & 80.2            & 35.2            & 27.7           \\ \hline
		DRC \cite{zhong2020deep}               & 35.6            & 83.0            & 38.4            & 32.1           \\ \hline
		CC \cite{li2021contrastive}            & 43.1            & 85.9            & 44.5            & 34.0           \\ \hline
		\textbf{HTCN} & \textbf{46.5}  & \textbf{87.5}  & \textbf{49.4}  & \textbf{35.6}  \\ \hline
	\end{tabular}
\end{table*}

\begin{table*}[]
	\renewcommand\arraystretch{1}
	\centering
	\caption{The ACC(\%) scores by different clustering methods (The best score in each column is in \textbf{bold}).}
	\label{tab2.2}
	\begin{tabular}{|p{2.04cm}<{\centering}|p{1.8cm}<{\centering}|p{2cm}<{\centering}|p{2.36cm}<{\centering}|p{2.33cm}<{\centering}|}
		\hline
		Dataset      & CIFAR-100 & ImageNet-10 & ImageNet-Dogs & Tiny-ImageNet \\ \hline
		$K$-means \cite{macqueen1967some}      & 13.0            & 24.1           & 10.5            & 2.5            \\ \hline
		SC \cite{zelnik2004self}               & 13.6            & 27.4           & 11.1            & 2.2            \\ \hline
		AC \cite{gowda1978agglomerative}       & 13.8            & 24.2           & 13.9            & 2.7            \\ \hline
		NMF \cite{cai2009locality}             & 11.8            & 23.0           & 11.8            & 2.9            \\ \hline
		AE \cite{bengio2006greedy}             & 16.5            & 31.7           & 18.5            & 4.1            \\ \hline
		DAE \cite{vincent2010stacked}          & 15.1            & 30.4           & 19.0            & 3.9            \\ \hline
		DCGAN \cite{radford2015unsupervised}   & 15.3            & 34.6           & 17.4            & 4.1            \\ \hline
		DeCNN \cite{zeiler2010deconvolutional} & 13.3            & 31.3           & 17.5            & 3.5            \\ \hline
		VAE \cite{kingma2013auto}              & 15.2            & 33.4           & 17.9            & 3.6            \\ \hline
		JULE \cite{yang2016joint}              & 13.7            & 30.0           & 13.8            & 3.3            \\ \hline
		DEC \cite{xie2016unsupervised}         & 18.5            & 38.1           & 19.5            & 3.7            \\ \hline
		DAC \cite{chang2017deep}               & 23.8            & 52.7           & 27.5            & 6.6            \\ \hline
		DCCM \cite{wu2019deep}                 & 32.7            & 71.0           & 38.3            & 10.8           \\ \hline
		GATC \cite{niu2020gatcluster}    & 32.7            & 73.9           & 32.2            & -              \\ \hline
		PICA \cite{huang2020deep}              & 33.7            & 87.0           & 35.2            & 9.8            \\ \hline
		DRC \cite{zhong2020deep}               & 36.7            & 88.4           & 38.9            & 13.9           \\ \hline
		CC \cite{li2021contrastive}            & 42.9            & 89.3           & 42.9            & 14.0           \\ \hline
		\textbf{HTCN} & \textbf{47.2}  & \textbf{90.5}  & \textbf{49.3}  & \textbf{16.0}  \\ \hline
	\end{tabular}
\end{table*}

\begin{table*}[]
	\renewcommand\arraystretch{1}
	\centering
	\caption{The ARI(\%) scores by different clustering methods (The best score in each column is in \textbf{bold}).}
	\label{tab2.3}
	\begin{tabular}{|p{2.04cm}<{\centering}|p{1.8cm}<{\centering}|p{2cm}<{\centering}|p{2.36cm}<{\centering}|p{2.33cm}<{\centering}|}
		\hline
		Dataset      & CIFAR-100 & ImageNet-10 & ImageNet-Dogs & Tiny-ImageNet \\ \hline
		$K$-means \cite{macqueen1967some}      & 2.8           & 5.7           & 2.0           & 0.5          \\ \hline
		SC \cite{zelnik2004self}               & 2.2           & 7.6           & 1.3           & 0.4          \\ \hline
		AC \cite{gowda1978agglomerative}       & 3.4           & 6.7           & 2.1           & 0.5          \\ \hline
		NMF \cite{cai2009locality}             & 2.6           & 6.5           & 1.6           & 0.5          \\ \hline
		AE \cite{bengio2006greedy}             & 4.8           & 15.2          & 7.3           & 0.7          \\ \hline
		DAE \cite{vincent2010stacked}          & 4.6           & 13.8          & 7.8           & 0.7          \\ \hline
		DCGAN \cite{radford2015unsupervised}   & 4.5           & 15.7          & 7.8           & 0.7          \\ \hline
		DeCNN \cite{zeiler2010deconvolutional} & 3.8           & 14.2          & 7.3           & 0.6          \\ \hline
		VAE \cite{kingma2013auto}              & 4.0           & 16.8          & 7.9           & 0.6          \\ \hline
		JULE \cite{yang2016joint}              & 3.3           & 13.8          & 2.8           & 0.6          \\ \hline
		DEC \cite{xie2016unsupervised}         & 5.0           & 20.3          & 7.9           & 0.7          \\ \hline
		DAC \cite{chang2017deep}               & 8.8           & 30.2          & 11.1          & 1.7          \\ \hline
		DCCM \cite{wu2019deep}                 & 17.3          & 55.5          & 18.2          & 3.8          \\ \hline
		GATC \cite{niu2020gatcluster}    & 17.3          & 55.2          & 16.3          & -            \\ \hline
		PICA \cite{huang2020deep}              & 17.1          & 76.1          & 20.1          & 4.0          \\ \hline
		DRC \cite{zhong2020deep}               & 20.8          & 79.8          & 23.3          & 5.6          \\ \hline
		CC \cite{li2021contrastive}            & 26.6          & 82.2          & 27.4          & 7.1          \\ \hline
		\textbf{HTCN}  & \textbf{30.5}  & \textbf{83.9}  & \textbf{35.2}  & \textbf{7.6} \\ \hline
	\end{tabular}
\end{table*}

In this section, we compare the proposed method against four non-deep clustering methods, namely, $K$-means \cite{macqueen1967some}, Spectral Clustering (SC) \cite{zelnik2004self}, Agglomerative Clustering (AC) \cite{gowda1978agglomerative}, and Nonnegative Matrix Factorization (NMF) \cite{cai2009locality}, and thirteen deep clustering methods, namely, Auto-Encoder (AE) \cite{bengio2006greedy}, Denoising Auto-Encoder (DAE) \cite{vincent2010stacked}, Deep Convolutional Generative Adversarial Networks (DCGAN) \cite{radford2015unsupervised}, DeConvolutional Neural Networks (DeCNN) \cite{zeiler2010deconvolutional}, Aariational Auto-Encoder (VAE) \cite{kingma2013auto}, Joint Unsupervised LEarning (JULE) \cite{yang2016joint}, Deep Embedded Clustering (DEC) \cite{xie2016unsupervised}, Deep Adaptive Clustering (DAC) \cite{chang2017deep}, Deep Comprehensive Correlation Mining (DCCM) \cite{wu2019deep}, Gaussian ATtention Network for image Clustering (GATC) \cite{niu2020gatcluster}, PartItion Confidence mAximization (PICA) \cite{huang2020deep}, Deep Robust Clustering (DRC) \cite{zhong2020deep} and Contrastive Clustering (CC) \cite{li2021contrastive}.

As shown in Table~\ref{tab2.1}, \ref{tab2.2} and \ref{tab2.3}, our HTCN method achieves the best scores on all the four benchmark datasets w.r.t. NMI, ACC, and ARI. Notably, on the ImageNet-Dogs dataset, our HTCN method obtains NMI(\%),ACC(\%) and ARI(\%) scores of 49.4, 49.3, and 35.2, respectively, which significantly outperforms the second best method (i.e., CC) that obtains NMI(\%),ACC(\%) and ARI(\%) scores of 44.5, 42.9, and 27.4.
The experimental results in Table~\ref{tab2.1}, \ref{tab2.2} and \ref{tab2.3} confirm the advantageous clustering performance of HTCN over the baseline methods.



\subsection{Influence of the Tri-stream Architecture}

In the proposed framework, we present a tri-stream architecture which consists of two online networks and a target network. In this section, we test the influence of the three streams of networks. As shown in Table~\ref{tab:alb1}, using an online network and a target network leads to better clustering results than using two online networks, while using three streams of networks outperforms both variants of using two streams, which shows the benefits of the heterogeneous tri-stream architecture.

\subsection{Influence of Two Types of Contrastive losses}


In the section, we test the influence of the two types of contrastive losses, i.e., the instance-level contrastive loss and the cluster-level contrastive loss.
As shown in Table~\ref{tab:alb2}, training with both types of losses can lead to better clustering performance than training with only one of them, which confirm the joint contribution of the instance-level and cluster-level losses in the self-supervised training.

\subsection{Influence of the Asymmetric Settings}

\begin{table}[!t]\vskip 0.1 in
	\renewcommand\arraystretch{1.25}
	\centering
	\caption{The clustering performance of HTCN using different combinations of network architectures.}
	\label{tab:alb1}
	\begin{tabular}{p{5cm}p{1.cm}<{\centering}p{1.cm}<{\centering}p{1.cm}<{\centering}}
		\toprule
		Architecture    & NMI  & ACC  & ARI  \\ \hline
		Tri-stream architecture        & 46.5 & 47.2 & 30.5 \\ \hline
		Dual-stream (Online+Target)   & 42.2 & 42.5 & 26.8 \\
		Dual-stream (Online+Online)   & 39.9 & 40.2 & 24.4 \\ \bottomrule
	\end{tabular}
\end{table}

\begin{table}[!t]\vskip 0.1 in
	\renewcommand\arraystretch{1.25}
	\centering
	\caption{The clustering performance of HTCN using different loss functions.}
	\label{tab:alb2}
	\begin{tabular}{p{4.1cm}p{1.cm}<{\centering}p{1.cm}<{\centering}p{1.cm}<{\centering}}
		\toprule
		Loss function               & NMI  & ACC  & ARI  \\ \hline
		With instance and cluster losses                    & 46.5 & 47.2 & 30.5 \\ \hline
		With only instance loss     & 43.3 & 35.6 & 14.6 \\
		With only cluster loss      & 38.4 & 36.4 & 22.7 \\ \bottomrule
	\end{tabular}
\end{table}

\begin{table}[!t]\vskip 0.1 in
	\renewcommand\arraystretch{1.25}
	\centering
	\caption{The NMI(\%), ACC(\%), and ARI(\%) by HTCN removing different asymmetric settings.}
	\label{tab:alb3}
	\begin{tabular}{p{4cm}p{1.cm}<{\centering}p{1.cm}<{\centering}p{1.cm}<{\centering}}
		\toprule
		Asymmetric settings                    & NMI  & ACC  & ARI  \\ \hline
		HTCN                        & 46.5 & 47.2 & 30.5 \\ \hline
		No predictor                    & 40.9 & 41.2 & 25.0 \\
		No stop-gradient                & 39.2 & 39.3 & 23.7 \\ \bottomrule
	\end{tabular}\vskip 0.1 in
\end{table}

Two symmetry-breaking mechanisms are enforced between the online and target networks \cite{grill2020bootstrap}. First, an instance predictor is incorporated in each online network, which does not exist in the target network. Second, the so-called stop-gradient is incorporated in the target network, which indicates that this network is not updated using backpropagation. We test the influence of the asymmetric settings by removing one of the instance predictor and the stop-gradient. As shown in Table~\ref{tab:alb3}, training with both asymmetric settings leads to better performance than training with only one of them.

\subsection{Convergence Analysis}

\begin{figure}[]\skip 0.1 in
	\centering
	\subfigure[{\scriptsize CIFAR-100}]{
		\includegraphics[width=0.231\textwidth]{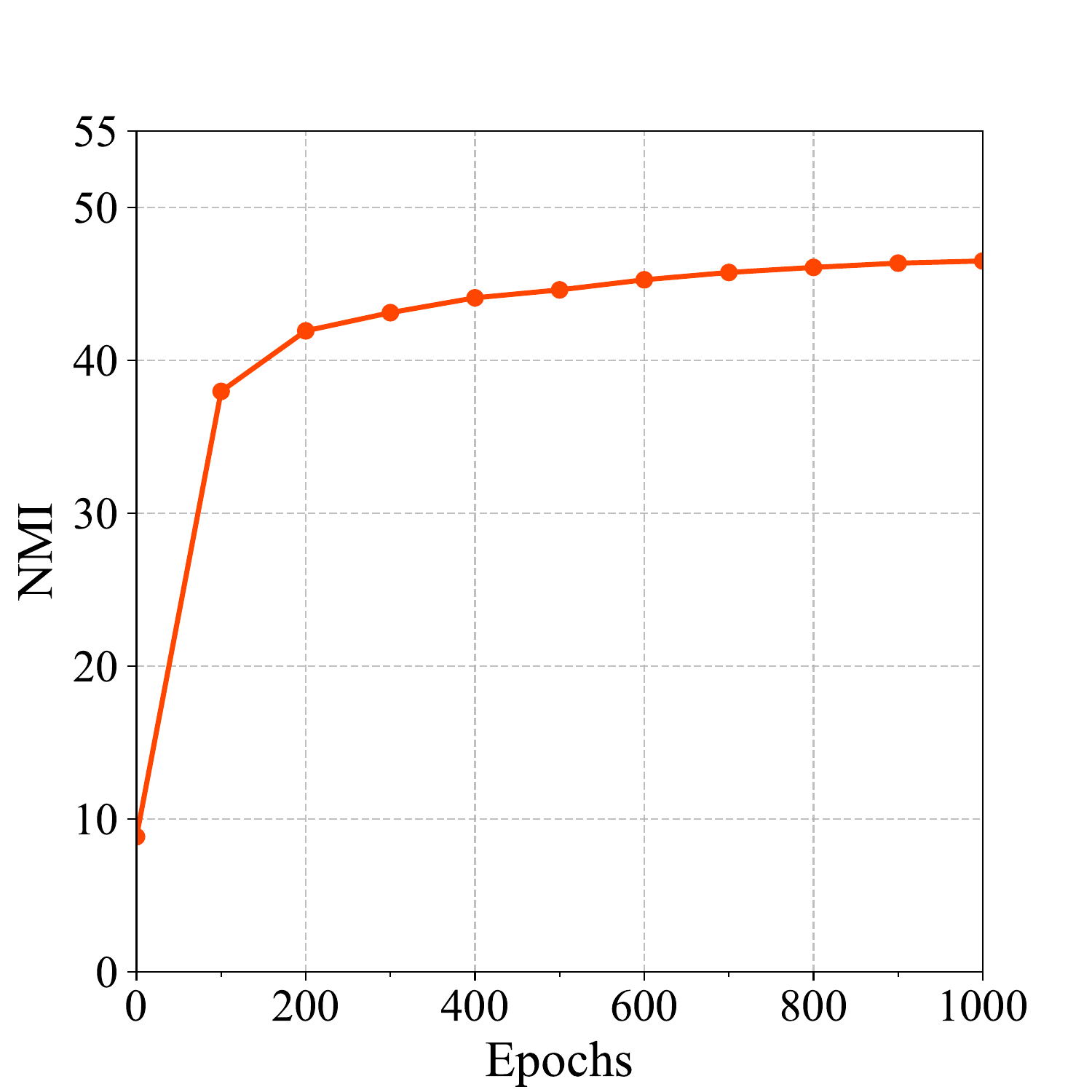}}
	\subfigure[{\scriptsize ImageNet-10}]{
		\includegraphics[width=0.231\textwidth]{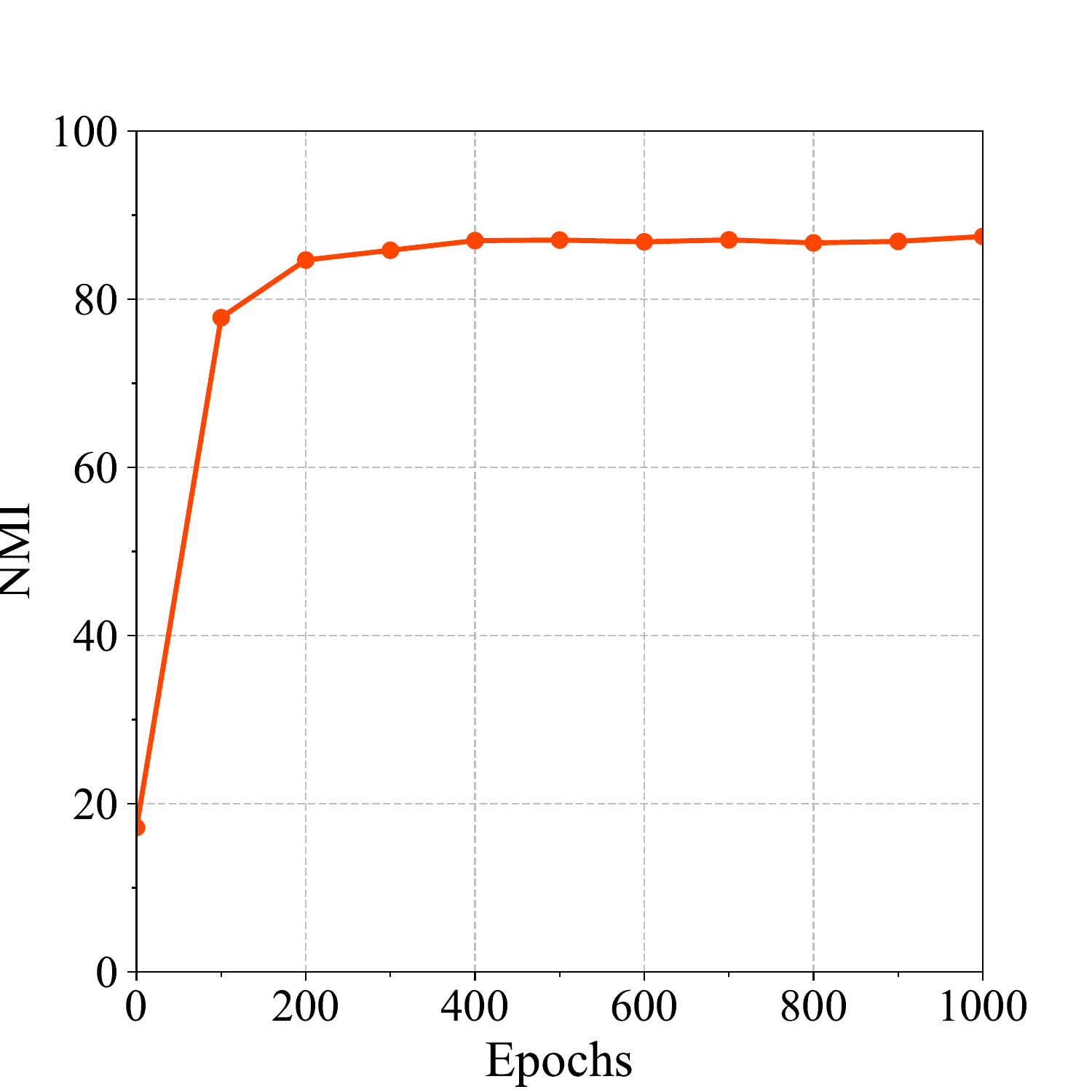}}
	\subfigure[{\scriptsize ImageNet-dogs}]{
		\includegraphics[width=0.231\textwidth]{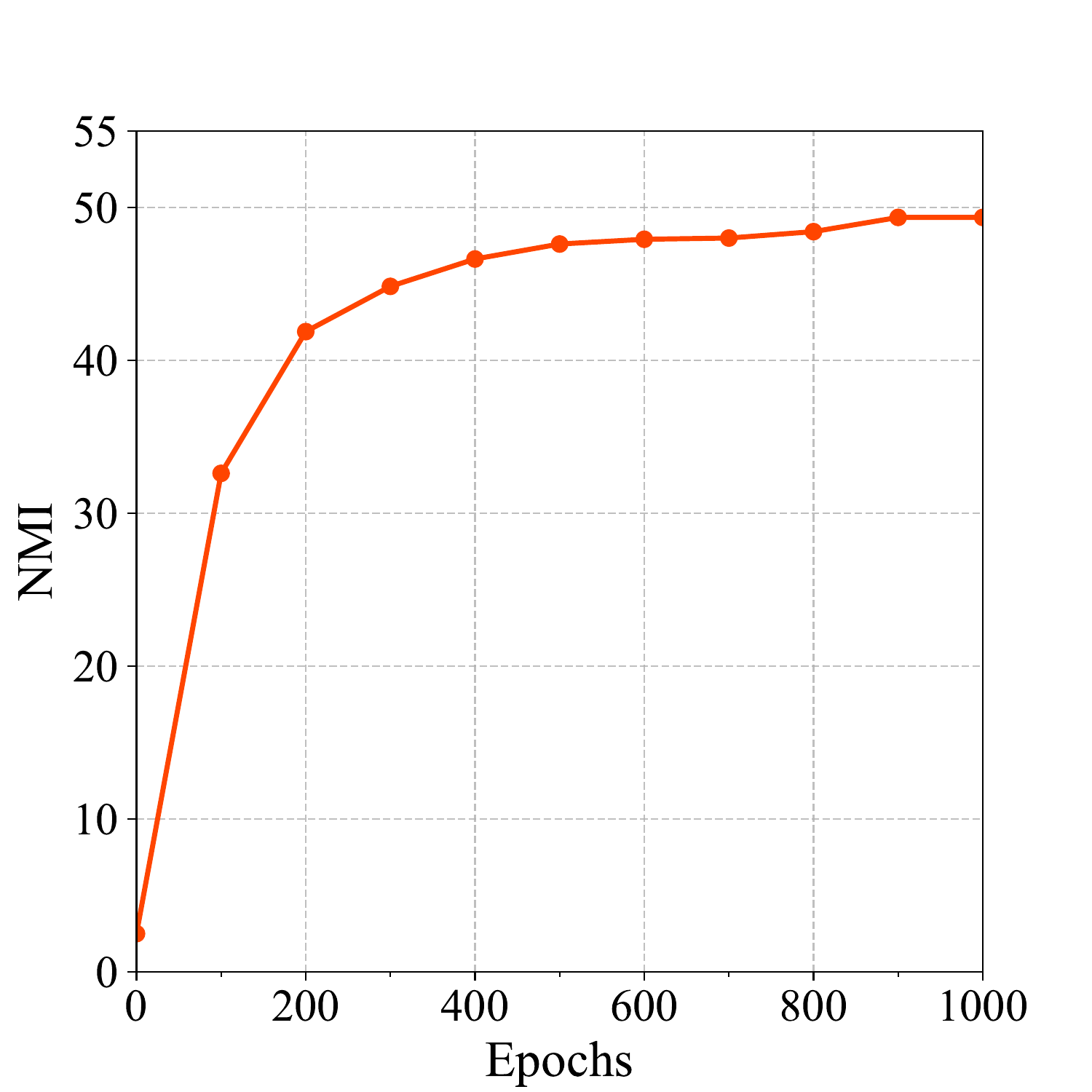}}
	\subfigure[{\scriptsize Tiny-ImageNet}]{
		\includegraphics[width=0.231\textwidth]{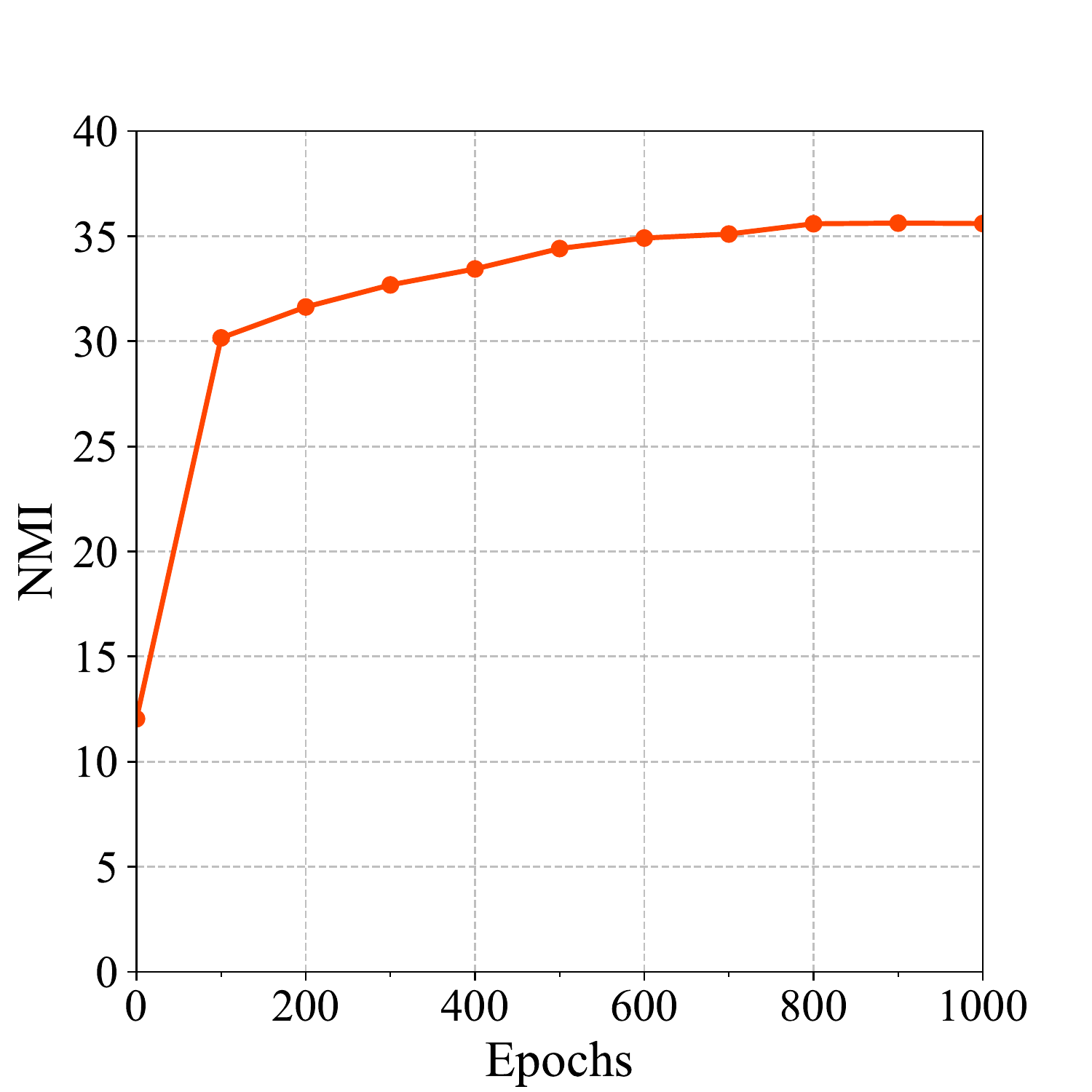}}
	\caption{Illustration of the convergence of HTCN (w.r.t. its NMI performance) on the four benchmark datasets.}
	\label{fig6}\vskip 0.1 in
\end{figure}

In this section, we test the convergence of the proposed HTCN method as the number of epochs increases. As shown in  Fig.~\ref{fig6}, the clustering scores (w.r.t. NMI) of the proposed HTCN method rapidly increase during the first 200 epochs on the benchmark datasets. When going beyond 200 epochs, the increase of epochs still benefits the clustering performance consistently. In this paper, the number of epochs is set to 1000 on all benchmark datasets.

\section{{Conclusion and Future Work}}\label{sec5}

The paper develops a new deep clustering approach termed HTCN, which breaks through the conventional two-stream contrastive architecture to explore the rich possibilities in heterogeneous  multi-stream contrastive learning and clustering. In HTCN, the two weight-sharing online networks are trained by predicting the instance representations of the target network and enforcing the consistency between the cluster representations of the target and online networks. Thus the tri-stream network architecture can be optimized in an end-to-end manner via simultaneous instance-level and cluster-level contrastive learning.
Experimental results on four challenging image datasets have shown the superior performance of our HTCN approach over the state-of-the-art deep clustering approaches. {In this paper, we mainly focus on the deep clustering task for images. In the future work, a possible direction is to extend the proposed framework to the deep clustering tasks for more complex data types, such as time series data and document data.}

\section*{Declarations}

\begin{itemize}
	\item \textbf{Funding.} This work was supported by the NSFC (61976097 \& 61876193) and the Natural Science Foundation of Guangdong Province (2021A1515012203).
	\item \textbf{Conflict of interest.} The authors declare that they have no conflict of interest.
	\item \textbf{Ethical approval.} This article does not contain any studies with human participants or animals performed by any of the authors.
	\item \textbf{Consent to participate.} Informed consent to participate was obtained from all individual participants included in the study.
	\item \textbf{Consent for publication.} Informed consent for publication was obtained from all individual participants included in the study.
	\item \textbf{Availability of data and materials.} All datasets used in this paper are publicly-available datasets.
	\item \textbf{Code availability.} The code is available at \url{https://github.com/dengxiaozhi/HTCN}.
	\item \textbf{Authors' contributions.} XD: Conceptualization, Methodology, Writing--Original Draft. DH: Conceptualization, Writing--Review \& Editing.
	CDW: Optimization, Writing--Review \& Editing.
\end{itemize}

\bibliography{refs}


\end{document}